\documentclass[11pt,a4paper]{article}
\usepackage[hyperref]{acl2020}
\usepackage{times}
\usepackage{graphicx}
\usepackage{graphics}
\usepackage{latexsym}

\usepackage{microtype}

\aclfinalcopy % Uncomment this line for the final submission
\title{SASS: Data and Methods for Subject Aware Sentence Simplification}

\author{Brad Windsor \thanks{All authors are equal contributors and NYU-affiliated.} \\
  \texttt{bw1879@nyu.edu} \\\And
  Luke Martin \\
  \texttt{lhm300@nyu.edu} \\ \\\And
  Anand Tyagi \\
  \texttt{anand.tyagi@nyu.edu} \\}

\date{}

\begin{document}
\maketitle
\begin{abstract}
Sentence simplification tends to focus on the generic simplification of sentences by making them more readable and easier to understand. This paper provides a dataset aimed at training models that perform subject aware sentence simplifications rather than simplifying sentences as a whole. We also test models on that dataset which are inspired by model architecture used in abstractive summarization.
We hand generated portions of the data and augment the dataset by further manipulating those hand written simplifications. Our results show that data-augmentation, data-masking, and model architecture choices used in summarization provide a solid baseline for comparison on subject aware simplification.

\end{abstract}

\section{Introduction}

Sentence simplification is a problem which aims to transform long, dense sentences into more accessible ones that are easier to understand. 
Current work in sentence simplification focuses on simplifying for the purpose of making sentences easier to understand. As a result, the output sentences tend to include as much of the information included in the original sentence as possible, altering the parts of the sentences which contain challenging vocabulary or excess words.

An alternative type of simplification which is relevant to explore is to simplify by topic. While making sentences easier to read is beneficial for helping people more easily understand the contents of various documents, simplifying by topic allows for different people to extract information directly useful to them at the sentence level.

Simplification by topic has currently been explored in the area of summarization, where models exist that allow for the creation of summaries tailored to a specific topic, as shown in \citet{inproceedings} and \citet{article}. However, in order to use the models currently available for topic specific summarization, the input documents must be typically greater than one sentence long; topic specific summarization is usually associated with multi-document summarization.

Current datasets used for evaluating sentence simplification models focused on the aforementioned goal of simplifying for the ease of understanding. In this paper, we present the SASS (Subject Aware Sentence Simplification) dataset which consists of sentences from a YELP dataset \cite{yelpdata}, simplified by one or more specified topics. This dataset can be used to test models specifically aiming to simplify sentences by topic, rather than just focused on ease of readability.

To come up with good candidates for the SASS dataset, we used the Spacy \cite{spacy} NER tagger to identify sentences which discussed our subjects of interest, and hand-wrote simplifications of those sentences. We augmented this dataset with additional entities mined from the corpus.

The original and augmented datasets are the first multi-topic sentence simplification datasets. After creating them, we used these datasets to study how well techniques for multi-topic summarization generalize to simplification. Our tests include encoder-decoder models following \citet{liu2019text} and the use of artificial tokens following \citet{scarton-specia-2018-learning}.

\section{Related Work}
Our work draws on earlier approaches to simplification, including the control of the degree of simplification, and on Subject-Aware problems in abstractive summarization.
\subsection{Simplification process}
Simplification is often a multi-step process with more than one model involved \cite{zhu2010monolingual}.
Some of the common steps in sentence simplification:
\begin{itemize}
    \item Splitting long sentences into shorter ones
    \item Dropping irrelevant information
    \item Replacing words or phrases
\end{itemize}

Sentence simplification models tend to either process information in several separate pipeline stages \cite{xu2015problems}, or solve the full problem in an end-to-end neural model \cite{zhang2017sentence}.

\subsection{Controllable Sentence Simplification}

Controllable sentence simplification is a new paradigm which aims to better control the degree of information omitted. Some examples:

\begin{itemize}
    \item Separating the Newsela corpus by grade level, and using an initial token in a seq2seq model to signify the target grade level \cite{scarton-specia-2018-learning}
    \item Training to produce a given compression ratio, degree of paraphrasing, or lexical complexity \cite{martin2019controllable}
\end{itemize}

\begin{table*}[ht]
    \centering
    \begin{tabular}{|p{3cm}|p{14cm}|}
    \hline
    Source Sentence      & Given the growing popularity of Indian cuisine, I am surprised that the Bombay Grill conglomerate (Green Street location, First Street location, Bombay Bazaar) have such a monopoly on Indian food in this town. \\ \hline
    Cuisine simplified & Bombay Grill and Bombay Bazaar are Indian restaurants.                \\ \hline
    Location simplified & Bombay Grill is on Green Street and Bombay Bazaar is on First Street.                              \\ \hline
    \end{tabular}
    \caption{Subject-aware sentence simplification}
    \label{tab:my_label}
\end{table*}
\begin{table*}[]
\begin{tabular}{|p{3cm}|p{14cm}|}
\hline
Source Sentence   & Out of all the Vietnamese spots in North Texas that I’ve tried, my absolute favorite is Pho Paseur in Arlington \\ \hline
Data Augmentation & Out of all the Islamic spots in the Gold Coast that I’ve tried, my absolute favorite is Taj Mahal in Arlington  \\ \hline
Data Masking      & Out of all the NORP0 spots in LOC0 that I’ve tried, my absolute favorite is ORG0 in LOC1                        \\ \hline
\end{tabular}
\caption{Data augmentation and masking}
\label{table:data_aug}
\end{table*}

\subsection{Subject-Aware Summarization}

Subject-Aware summarization aims to condense a document but tailor it for a specific purpose. One such problem is Topic-Aware Abstractive Text summarisation \cite{zheng2020topicaware}, which attempts to leverage the underlying semantic structure of documents represented by their latent topics. In \citet{fan2018controllable}, the authors propose methods that would allow a user to restrict  the length of the summary, ask for entity-specific or source-specific summaries, and only summarise specific portions of the text. \citet{wang2020reinforced} mines topic-specific words from using topic-modelling on a large corpus and uses these words as an input to an attention mechanism for used in summary generation. 
Finally, there have been other attempts to incorporate more information into the summarizing to tailor the information to specific requests \cite{baumel2018query}.

\section{Methods\footnote{Code and data are available at https://github.com/bwindsor22/sentence-simp-target}}

\subsection{Setup}
\subsubsection{Data Preparation}

For ease of understanding, we choose the Yelp Reviews Dataset as our base \cite{yelpdata}.
We use Spacy \cite{spacy} to pre-identify 1500 relevant sentences which had entities marked Organization (ORG), Nationality (NORP), and Location (LOC). From this, we denote a simplification which includes ORG/NORP as a "Culinary" simplification and ORG/LOC as a "Location" simplification.

From this, we hand-annotated 599 example sentences which were simplified into multiple sentences based on tagging the in each sentence, we identified sentences that included two or more of any individual entities. We then simplified each given sentence into two or more sentences with each sentence containing exactly one or more of the entities relevant to the topic.

Because sentences where both simplifications are possible are rare, our corpus includes sentences where only one of the two simplifications is possible. See Table 3 for summary of annotation process.
\newline
\newline
\textbf{Augmenting data}
\newline
To increase the volume of the data, we augmented the data by mining entity names from the remaining Yelp dataset and substituting into the annotated sentences. First a Spacy tagger was run over the dataset, extracting out all the entities and their corresponding tags. Next, for each row in the dataset, where a row consisted of the source sentence and its summaries, the entities in those sentences were replaced with ones sampled from elsewhere in the dataset. This created the same sentence but with different entities. 

Using the example presented in Table \ref{table:data_aug}, an ORG, NORP, LOC tuple mined from one sentence is inserted into another. This approach is inspired by \cite{wang2020reinforced}'s keyword mining. We used this strategy to vary our data increase from 2 times to 6 times the amount of the original hand annotated data.
\newline
\newline
\textbf{Masking data}
\newline
We use the Spacy model to replace specific entities ("Islamic") with tags ("NORP0").  The Spacy model is an example of a task-specific NER system for our dataset; we selected organizations, nationalities, and locations in part because Spacy understands these well.

\begin{table}[]
\begin{tabular}{|l|l|}
\hline
Total Annotated                      & 599    \\ \hline
\hspace{3mm} Culinary \& Location Simplifications & 261   \\ \hline
\hspace{3mm} Culinary Simplifications Only        & 249   \\ \hline
\hspace{3mm} Location Simplifications Only        & 49    \\ \hline
\end{tabular}
\caption{Raw data counts from \textasciitilde 1500 initial  candidate sentences (\textasciitilde 900 were marked ineligible/off topic)}
\end{table}

\subsubsection{Model}
\begin{figure}[h]
\includegraphics[scale=0.5]{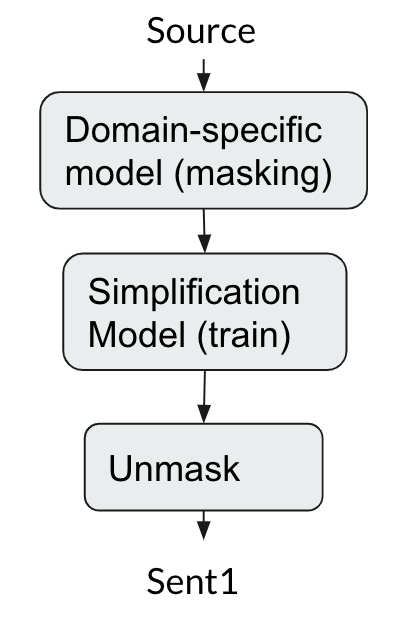}
\caption{Pipeline for masking. A model replaces key terms with generic tags.}
\end{figure}

\begin{figure}[h]
\includegraphics[scale=0.5]{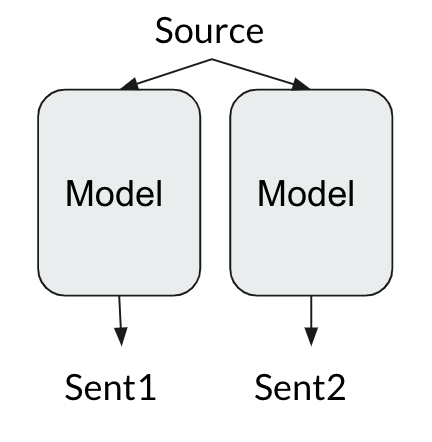}
\caption{Two models, trained separately, each task-specific}
\end{figure}
\begin{figure}[h]
\includegraphics[scale=0.5]{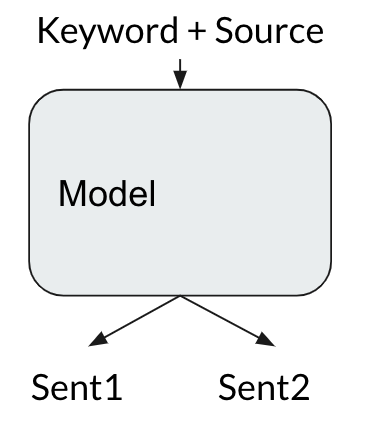}
\caption{One model using an artificial keyword to specify the style of output}
\end{figure}

For training, we use an encoder-decoder architecture sequence generation, inspired by \cite{liu2019text}'s work in summarization. We use Roberta \cite{liu2019roberta} as our base model and the HuggingFace Transformers \cite{wolf-etal-2020-transformers} library for encoder-decoder implementation. Our hyperparameters are: batch size: 8, train epochs: 200, learning rate: 0.001.

As simplification is often a multi-model process \cite{zhu2010monolingual}, one of our data preparation techniques included using a task-specific model to mask, as in Fig. 1.

\begin{table*}[]
\begin{tabular}{|l|l|l|}
\hline
Dataset                & Culinary    & Location         \\ \hline
Original               & 0.28 & 0.26 \\ \hline
Masked                 & 0.48 & 0.43 \\ \hline
Data Augmentation - 2X (100 epochs) & 0.29  & 0.29 \\ \hline
Data Augmentation - 6X (59 epochs) & 0.30  & 0.30 \\ \hline
\end{tabular}
\caption{Model BLEU scores on datasets. Two scores are reported, one for each task (as in Fig 2.)}

\end{table*}

\begin{table*}[]
\begin{tabular}{|l|l|}
\hline
Model                                            & Avg BLEU    \\ \hline
Two models, trained for separate tasks, original dataset (Fig 2)           & 0.28 \\ \hline
One model, with artificial token to specify task (Fig 3) & 0.43 \\ \hline
\end{tabular}
\caption{Results of model architecture analysis. BLEU score is weighted average of both tasks.}
\end{table*}

For training, we explore two model architectures per Fig. 2 and 3. In the first, two models are trained for two different tasks, with zero knowledge share between the two. In the second, we explore the use of task-specific tokens to specify the simplification style, following \citet{scarton-specia-2018-learning}.

\subsubsection{Evaluation}
Results are taken as average BLEU scores when compared to the target sentence \cite{papineni-etal-2002-bleu}, using the NLTK version of BLEU.

\subsection{Results}
Results are seen in Tables 4 and 5.

\subsection{Analysis}

We note the following observations from the training:

\begin{itemize}
    \item Models which have subject-specific expertise can significantly improve performance on the topic specific simplification task. This is seen in the data masking performance, where simplified sentences frequently took forms such as "ORG0 is a NORP0 restaurant". By specifying the information the model should be interested in as ORG0/ORG1/etc. tags, we were able to allow the model to focus on the work of restructuring the sentence, rather than just finding the correct entities. 
    \item Data masking works better for culinary simplifications, but we feel this is due to disagreement between annotator and the Spacy model: "North America", "2nd Floor Atrium" are examples of what annotators label as locations and Spacy does not.
    \item There are similar BLEU scores for both the Culinary and Location datasets. The degree of effectiveness is determined by the technique, rather than the subject. This raises hope that our methods would generalize well to other subjects.
    \item Mining subject-specific entities from a larger corpus can also be used to improve the sentence simplification task. Substituting various different restaurant names and cuisines during the data augmentation helped proved useful for helping the model learn.
    \item Data augmentation is a less useful preparation than data masking. The model does not as easily learn that "Iranian" and "Greek" play the same roles as it does in a dataset where both are labelled "ORG0". Data masking helps the model generalize better across examples, however, we note that sufficient data augmentation approaches the performance of masking.
    \item The paradigm of task-specific tokens successful in tailoring simplifications to a reading level \cite{scarton-specia-2018-learning} works well with subject-specific simplification also. One model learning both for the original dataset outperforms two models. This is unsurprising in that our data is very scarce ($<$1K examples in each category). Allowing the model to see unrelated examples is similar to pre-training on the domain-specific text.
    
\end{itemize}

\section{Conclusion}

In this paper, we introduced the SASS dataset which allows for models focused on topic specific sentence simplification to be evaluated.  The evaluation of our baseline enconder-decoder model showed that even with a simple model, we are able to generate a system which can perform simplification based on a specified topic.

We presented a new dataset for this challenge, found proof that augmentation techniques help in this domain, and proved that existing strategies in  summarization also apply in this domain. Our results help clarify a new approach to simplification, and shed light on how well techniques from other problems generalize.

Further work can be done to expand on the dataset we introduced, both manually tagging more sentences with a wider range of tags, and developing methods to automatically augment the dataset. We expect that paraphrasers such as Quillbot \cite{quillbot} could futher augment sentences, or that entity lists such as YAGO \cite{yago} could be used to further fill out sentences.

Performing subject aware sentence simplification allows for texts to be simplified not only in a manner that can be more understood, but simplified in a way that is relevant to the individual. We hope that the SASS dataset can be used to evaluate new models created specifically for this purpose and further improve the overall area of sentence simplification.

\bibliography{anthology,acl2020}
\bibliographystyle{acl_natbib}

\end{document}